\documentclass{article}

\PassOptionsToPackage{numbers, compress}{natbib}

     \usepackage[final]{neurips_2019}

\usepackage[utf8]{inputenc} 
\usepackage[T1]{fontenc}    
\usepackage{hyperref}       
\usepackage{url}            
\usepackage{booktabs}       
\usepackage{amsfonts}       
\usepackage{nicefrac}       
\usepackage{microtype}      

\usepackage{tabularx}       
\usepackage{graphicx}       
\usepackage{amsmath}        
\usepackage[labelfont=bf]{caption}
\title{Improved hierarchical patient classification with language model pretraining over clinical notes}

\author{%
  Jonas Kemp\thanks{Work completed in part during the Google AI Residency.} \qquad Alvin Rajkomar \qquad Andrew M. Dai \\
  Google Health \\
  \texttt{\{jonasbkemp,alvinrajkomar,adai\}@google.com} \\
}

\begin{document}

\maketitle

\begin{abstract}
Clinical notes in electronic health records contain highly heterogeneous writing styles, including non-standard terminology or abbreviations. Using these notes in predictive modeling has traditionally required preprocessing (e.g. taking frequent terms or topic modeling) that removes much of the richness of the source data. We propose a pretrained hierarchical recurrent neural network model that parses minimally processed clinical notes in an intuitive fashion, and show that it improves performance for discharge diagnosis classification tasks on the Medical Information Mart for Intensive Care III (MIMIC-III) dataset, compared to models that treat the notes as an unordered collection of terms or that conduct no pretraining. We also apply an attribution technique to examples to identify the words that the model uses to make its prediction, and show the importance of the words' nearby context. 
\end{abstract}

\section{Introduction}

With the rapid deployment of electronic health records (EHRs) in the US, clinicians routinely enter patient data electronically, mostly in unstructured, free-text clinical notes. Because clinicians frequently employ non-standard, ambiguous shorthand phrases or organize their notes in unpredictable ways, automated parsing for downstream use can be quite challenging. Traditional natural language processing (NLP) techniques relying on hand-crafted rules \citep{Taggart2018-wg} or feature engineering can be difficult to apply in this context. In practice, machine learning models tend to make more use of structured fields such as medications and diagnoses that can be straightforwardly extracted from the EHR \citep{10.1093/jamia/ocw042}, and clinical notes are often ignored outright \citep{2015arXiv151103677L, Choi2016-zr, Esteban2016-gn, Nickerson2016-my, Pham2016-ny, Choi2017-iz, Che2018-vj, Cheng2016-yw, 7762861}. Models that do use notes frequently reduce them to an unordered set of words \citep{Marafino2018-wn, Jacobson2016-ux, Rajkomar2018-jv} or topics \citep{Miotto2016-gr, Suresh2017-te}, which ignores many subtleties of language and context and can therefore obscure the meaning of the note.

Recent advances in deep learning have led to major improvements in a wide variety of NLP applications \citep{Johnson2017-jc, 2018arXiv181004805D}. Building on this work, we propose a model employing \textit{sequential, hierarchical, and pretraining (SHiP)} techniques from deep NLP to improve EHR predictive models by automatically learning to extract relevant information from clinical notes. Specifically, our model employs a hierarchical attention network \citep{Yang2016-dm}, augmented with a language model pretraining objective \citep{Dai2015-sl}, to read notes with minimal asssumptions about the text. We evaluate our model on standard classification tasks for EHRs, and compare performance against existing state-of-the-art baselines \citep{Rajkomar2018-jv}. We also evaluate the sensitivity of the model's outputs to different phrases in the text using deep learning attribution methods \citep{Sundararajan:2017:AAD:3305890.3306024}. To our knowledge, the effectiveness of language model pretraining has not been previously demonstrated for hierarchical classification models.

\section{Methods}

\subsection{Dataset and Prediction Tasks}

We developed our models using critical care data from the Medical Information Mart for Intensive Care (MIMIC-III) \citep{Johnson2016-is, Pollard2016-tm}. We represented patients' medical histories as a time series according to the Fast Healthcare Interoperability Resources (FHIR) specification, as described in previous work \citep{Rajkomar2018-jv}. The study cohort included all patients in MIMIC-III hospitalized for at least 24 hours. (See supplementary table \ref{table1} for cohort summary statistics.) From these records, we extracted basic encounter information (admission type, status, and source), diagnosis and procedure codes, medication orders, quantitative observations (lab results and vital signs), and free-text clinical notes. For each continuous feature, we standardized values to Z-scores using training set statistics, with any outliers more than 10 standard deviations from the mean capped to a score of $\pm 10$. For each hospitalization, we developed models for the following classification tasks, using the patient's full history up to the specified time in the current admission (including all past hospitalizations):
\begin{itemize}
    \item Inpatient mortality prediction (predicted 24 hours after admission).
    \item Primary CCS \citep{Elixhauser1996-ff} discharge diagnosis code (predicted at the moment of discharge).
    \item All ICD-9 \citep{Slee1978-xi} discharge diagnosis codes (predicted at the moment of discharge).
\end{itemize}

\subsection{Model Architecture}

We built on a core embedding scheme and top-level LSTM architecture described in previous work \citep{Rajkomar2018-jv}. In this framework, we embedded discrete features from the patient record (e.g. diagnosis codes) and trained these jointly with the model. To reduce sequence length, we grouped observations into fixed-length timesteps, or ``bags,'' and averaged all embeddings or continuous values for observations of the same feature within the same bag; additionally, we collapsed all observations occurring prior to the most recent $t$ timesteps into a single bag (with bag duration and $t$ tuned as hyperparameters). Finally, we concatenated the bagged embeddings or values for all features into a single representation of each timestep in the patient history, and we fed this embedded sequence into a long short-term memory (LSTM) network \citep{Hochreiter1997-wc}, generating predictions from the final hidden state.

In the standard bag-of-words (BOW) approach to these models, notes are treated just as any other discrete feature, with individual words embedded and aggregated without regard to ordering. Our SHiP models augmented this approach in two ways. First, we maintained the sequential order of embeddings within each note and fed these to a second LSTM to generate a context-sensitive representation for each word. We computed the final output vector for each note by applying hierarchical dot product attention \citep{Yang2016-dm} over this output sequence, placing higher weight on the portions of the notes most important for downstream prediction. Second, we used unsupervised language model pretraining \citep{Dai2015-sl} to pretrain the notes LSTM: before optimizing the prediction loss, we trained an auxiliary objective such that, for each word in the note, the LSTM learned to predict the next word (and, if bidirectional, the previous word).

In addition to the core BOW and SHiP models, we also compared several variants of the above, including: a model without notes; models using only notes; a BOW model with unigram and bigram embeddings; and hierarchical attention models without pretraining.

\subsection{Attribution Methods}
To compute attribution scores over the text of notes, we used the path-integrated gradients technique \citep{Sundararajan:2017:AAD:3305890.3306024}. For clarity in these attributions, we ran a notes-only model over only the selected note, omitting the rest of the notes in the patient's record. We computed attribution scores with respect to each word embedding, relative to a zero-vector baseline, using $m = 20$ steps to approximate the path integral.

\section{Results}

\subsection{Training and Evaluation Approach} 

We split our cohort by patient ID into 80\% train, 10\% validation, and 10\% test splits. Models were optimized using Adam \citep{DBLP:journals/corr/KingmaB14}, and regularized using dropout \citep{Srivastava2014-ws, Gal2016-yo} and Zoneout \citep{2016arXiv160601305K}. We used a Gaussian process bandit optimization algorithm \citep{JMLR:v15:desautels14a} to select hyperparameters maximizing performance for each task on the validation set. (See supplementary material B, particularly Table \ref{hparams}, for additional details.) Following hyperparameter tuning, we report mean (standard deviation) test set metrics over five runs from random initialization. Where reported, we also compute the statistical significance of pairwise differences in models' performance using a two-tailed Welch's t-test.

\subsection{Model Performance}

\begin{table}[htbp]
  \caption{Model performance results on the tasks of interest. Best values for each metric are bolded.} 
  \begin{tabularx}{\textwidth}{p{2cm} >{\hangindent=0.5em}p{1.9cm} X X X X X X} \toprule
     \multicolumn{2}{c}{Model} & \multicolumn{2}{c}{Mortality} & \multicolumn{2}{c}{Primary CCS} & \multicolumn{2}{c}{All ICD-9} \\ \cmidrule{3-8}
     & & AUPRC & AUROC & Top-1 Recall & Top-5 Recall & AUPRC & \mbox{AUROC}, weighted \\ \midrule
     No notes & - & 0.449 (0.006) & 0.869 (0.001) & 0.526 (0.006) & 0.796 (0.003) & 0.305 (0.001) & 0.873 (<0.001) \\ \midrule
     Bag-of-words & Unigrams (notes only) & 0.383 (0.004) & 0.832 (0.003) & 0.591 (0.004) & 0.849 (0.002) & 0.328 (0.002) & 0.880 (0.001) \\
     & Unigrams (all features) & \textbf{0.479 (0.008)} & 0.880 (0.001) & 0.592 (0.003) & 0.842 (0.003) & 0.331 (0.001) & 0.883 (0.001) \\
     & Unigrams and bigrams (all features) & 0.460 (0.005) & 0.872 (0.002) & 0.587 (0.008) & 0.829 (0.005) & 0.325 (0.002) & 0.881 (<0.001) \\ \midrule
     Hierarchical (without pretraining) & Notes only & 0.351 (0.003) & 0.825 (0.003) & 0.606 (0.003) & 0.850 (0.001) & 0.345 (0.005) & 0.887 (0.002) \\
      & All features & 0.471 (0.006) & 0.876 (0.003) & 0.591 (0.008) & 0.833 (0.006) & 0.301 (0.004) & 0.868 (0.001) \\ \midrule
     SHiP & Notes only & 0.353 (0.005) & 0.825 (0.004) & 0.667 (0.006) & \textbf{0.897*$^\dagger$ (0.003)} & \textbf{0.352$^\dagger$ (0.001)} & \textbf{0.891$^\dagger$ (0.001)} \\
     & All features & \textbf{0.479 (0.007)} & \textbf{0.882 (0.001)} & \textbf{0.671*$^\dagger$ (0.004)} & 0.890 (0.001) & 0.345 (0.005) & 0.889 (0.002) \\ \bottomrule
  \end{tabularx}
  \caption*{* $p < 0.001$ for difference compared to corresponding hierarchical model without pretraining. \newline $^\dagger$ $p < 0.001$ for difference compared to best bag-of-words model.}
  \label{results} 
\end{table}

Table \ref{results} compares the performance of all model variants. The SHiP models significantly improved over the BOW baselines on the two diagnosis tasks ($p < 0.001$ under Welch's t-test): for CCS prediction, the best SHiP models improved top-1 recall by 7.9 percentage points and top-5 recall by 4.8 percentage points, respectively, over the best BOW models; for ICD-9 prediction, area under the precision-recall curve (AUPRC) increased by 2.1 percentage points and weighted area under the ROC curve (AUROC) increased by 0.8 percentage points. For mortality prediction, we saw negligible benefit from the SHiP architecture.

The SHiP models also improved over the corresponding hierarchical models without pretraining. For mortality, pretraining the all-features model increased AUPRC by 0.8 percentage points ($p = 0.06$) and AUROC by 0.6 percentage points ($p = 0.004$); for primary CCS, pretraining the all-feature model increased top-1 recall by 8.0 percentage points ($p < 0.001$), while pretraining the notes-only model increased top-5 recall by 4.7 percentage points ($p < 0.001$); for all ICD-9, pretraining the notes-only model increased AUPRC by 0.7 percentage points ($p = 0.03$) and weighted AUROC by 0.4 percentage points ($p = 0.01$).

\subsection{Qualitative Analysis}
\begin{figure}[htbp]
  \centering 
  \includegraphics[width=0.8\textwidth]{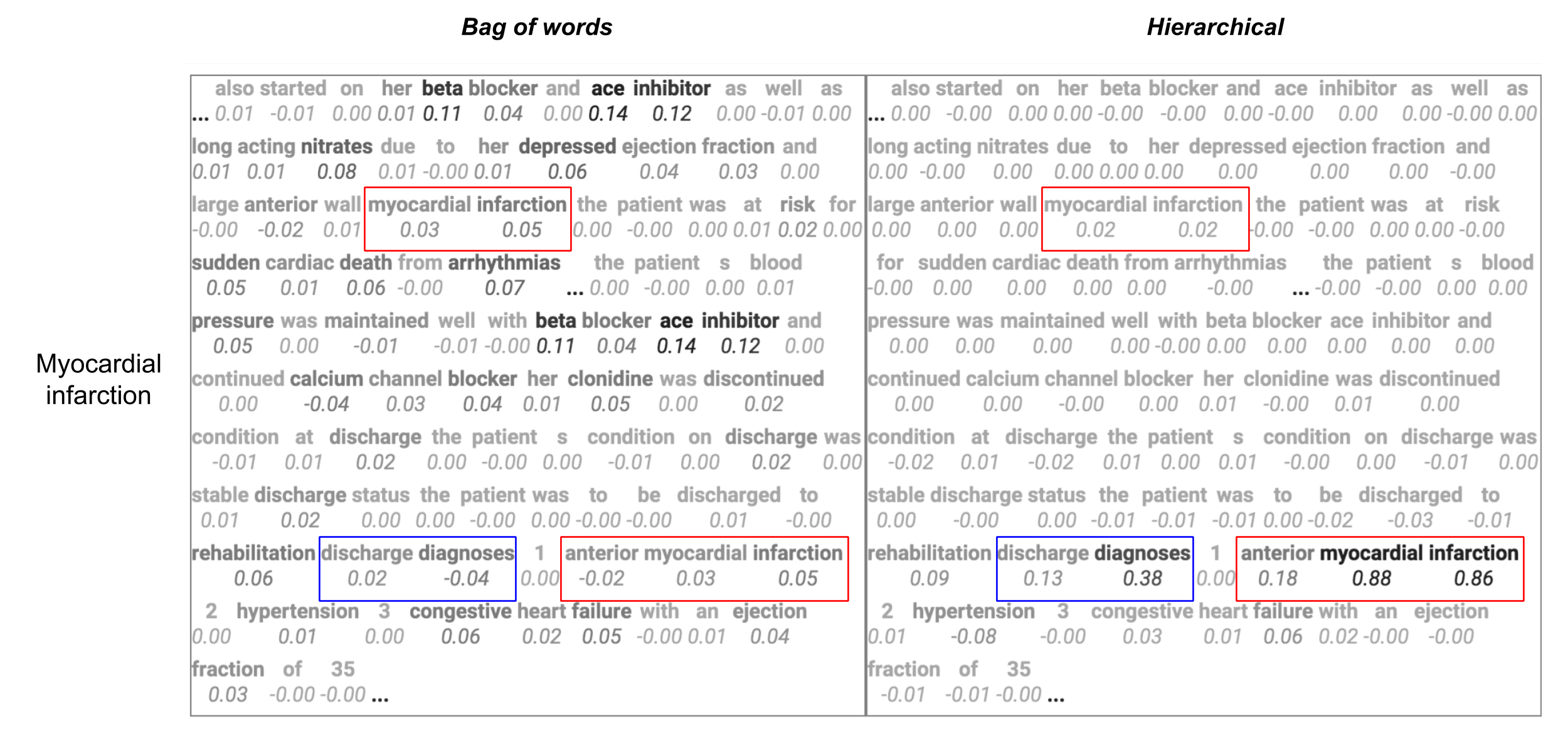}
  \includegraphics[width=0.8\textwidth]{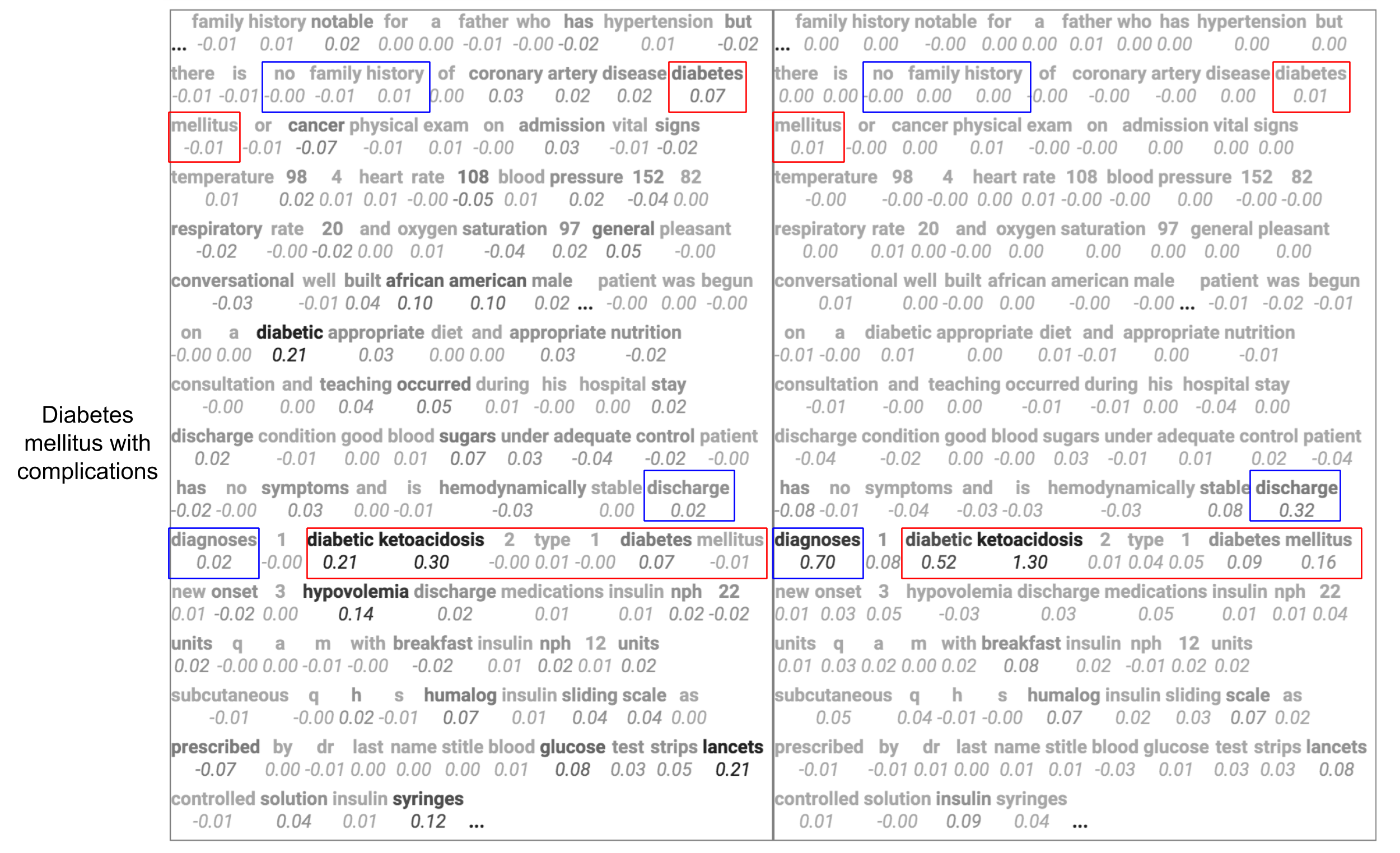}
  \caption{Visualization of integrated gradients attribution over excerpts from patient discharge summaries (primary diagnosis shown at left). For each excerpt, the left column shows attribution from the BOW baseline, and the right column shows attribution from the SHiP model. Below each word is the value of the attribution computed for that word, where a higher absolute value indicates greater importance. Red boxes highlight the patient's stated diagnosis in the text, while blue boxes indicate relevant pieces of nearby context.}
  \label{int-grads} 
\end{figure}

Figure \ref{int-grads} shows examples of path-integrated gradients attribution from CCS prediction models, over discharge summaries from different patients. We observe that the SHiP model frequently concentrates on just one or a few important phrases, even in very long notes. The choice of phrase is often informed by the nearby context: for example, we can see that the SHiP model is consistently most sensitive to the clinically-relevant words following the phrase ``discharge diagnoses.'' In fact, in each sample here, the patient's diagnosis is restated elsewhere in the text in a less relevant context (e.g. stating that the patient has ``no family history'' of diabetes), but the model is sensitive only to the instance where the discharge context is made explicit. The bag-of-words model, by contrast, is incapable of making such contextual distinctions, and is generally more sensitive to key words and phrases throughout the text.

\section{Conclusion} 

We demonstrate that SHiP, a novel combination of hierarchical modeling of clinical notes and language model pretraining, can improve discharge diagnosis classification over previous state-of-the-art models, with only minimal preprocessing of text. Our work builds on a substantial recent literature on applying deep learning techniques to analysis of electronic health records data \citep{Shickel2017-dy}, including many clinical NLP studies using more standard convolutional or recurrent architectures \citep{Jagannatha2016-ok, 2017arXiv170508557V, mullenbach-etal-2018-explainable, 10.1371/journal.pone.0192360, Chokwijitkul2018-ki}, or employing hierarchical models with limited or no pretraining \citep{Gao2017-go, 2017arXiv170909587B, Samonte2018-vx, 2018arXiv180804928L, Newman-Griffis2018-ke}. Drawing on the respective successes of hierarchical attention networks \citep{Yang2016-dm, 2016arXiv160901704C, 7953252, Yu2016-ik, Meng2017-uz} and pretraining methods \citep{2018arXiv181004805D, Dai2015-sl, 2019arXiv190608237Y} in a wide variety of general NLP applications, we show the utility of these methods applied jointly, and specifically within a clinical context.

\newpage
\section*{Acknowledgments}
We thank Nissan Hajaj and Xiaobing Liu for developing the core framework used to implement our models. We thank Gerardo Flores, Kathryn Rough, and Kun Zhang for providing assistance with our data processing and evaluation pipelines. We thank Kai Chen, Michael Howell, and Denny Zhou for their comments and feedback on this manuscript.

\section*{Code Availability}
A code sample illustrating our approach is available at \href{https://github.com/google-health/records-research/tree/master/clinical-notes-prediction}{https://github.com/google-health/records-research/tree/master/clinical-notes-prediction}.

\bibliographystyle{unsrtnat}
\bibliography{references}

\newpage
\appendix
\section*{Supplementary Material}

\section{Additional details of patient cohort}

\begin{table}[htbp]
  \caption{Descriptive statistics for patient cohort.} 
  \begin{tabularx}{\textwidth}{p{7cm} X X} \toprule
  & Train \& validation & Test \\ \midrule
  Number of patients & 40,511 & 4,439 \\
  Number of hospital admissions* & 51,081 & 5,598 \\
  \quad Gender, $n (\%)$ & & \\
  \qquad Female & 22,468 (44.0) & 2,548 (45.5) \\
  \qquad Male & 28,613 (56.0) & 3,050 (54.5) \\
  \quad Age, median (IQR) & 62 (32) & 62 (33) \\
  \quad Hospital discharge service, $n$ $(\%)$ & & \\
  \qquad General medicine & 21,350 (41.8) & 2,354 (42.1) \\
  \qquad Cardiovascular & 10,965 (21.5) & 1,175 (21.0) \\
  \qquad Obstetrics & 7,123 (13.9) & 803 (14.3) \\
  \qquad Cardiopulmonary & 4,459 (8.7) & 519 (9.3) \\
  \qquad Neurology & 4,282 (8.4) & 457 (8.2) \\
  \qquad Cancer & 2,217 (4.3) & 223 (4.0) \\
  \qquad Psychiatric & 28 (0.1) & 4 (0.1) \\
  \qquad Other & 657 (1.3) & 63 (1.1) \\
  \quad Discharge location, $n$ $(\%)$ & & \\
  \qquad Home & 28,991 (56.8) & 3,095 (55.3) \\
  \qquad Skilled nursing facility & 6,878 (13.5) & 794 (14.2) \\
  \qquad Rehab & 5,757 (11.3) & 653 (11.7) \\
  \qquad Other healthcare facility & 3,830 (7.5) & 448 (8.0) \\
  \qquad Expired & 4,420 (8.7) & 462 (8.3) \\
  \qquad Other & 1,205 (2.4) & 146 (2.6) \\
  \quad Previous hospitalizations, $n$ $(\%)$ & & \\
  \qquad None & 40,362 (79.0) & 4,415 (78.9) \\
  \qquad One & 6,427 (12.6) & 721 (12.9) \\
  \qquad Two to five & 3,681 (7.2) & 397 (7.1) \\
  \qquad Six or more & 611 (1.2) & 65 (1.2) \\
  \quad Number of discharge ICD-9, median (IQR)** & 9 (8) & 9 (8) \\ \bottomrule
  \end{tabularx}
  \caption*{* For primary CCS prediction, 1.3\% of these admissions were excluded, where the primary diagnosis corresponded to a non-billable ICD-9 code. \\ ** Includes only billable ICD-9 codes.} 
  \label{table1} 
\end{table}

\newpage
\section{Additional details of model training}

For memory and performance reasons, in all hierarchical models we restricted the maximum amount of text used in the notes LSTM, keeping the most recent $N$ tokens per record (across all notes) and discarding any additional leading tokens. We tuned the level of truncation on the validation set, and found $N = 1000$ to be sufficient for training mortality models, but increased to $N = 2500$ for both diagnosis tasks and for pretraining. All models were implemented in Tensorflow 1.12 \citep{2016arXiv160304467A}, and trained on Nvidia Tesla P100 GPUs.

Metrics for model selection included AUROC for mortality and ICD9, and top-5 recall for CCS. For multilabel ICD9 prediction, we computed a weighted AUROC, where the AUROC for each label is averaged according to the label's prevalence. Evaluation metrics and statistical tests were calculated using scikit-learn 0.20 \citep{Pedregosa2011-ss}.

\begin{table}[htbp]
  \caption{Model hyperparameters. For the same task, all non-hierarchical models shared the BOW hyperparameters, and all hierarchical models shared the SHiP hyperparameters, except where noted. For the SHiP models, dropout was applied during both pretraining and training. All models were trained using the Adam optimizer with default constant values: $\beta_1=0.9, \beta_2=0.999, \epsilon=1 \times 10^{-8}$.} 
  \begin{tabularx}{\textwidth}{p{1.25cm} >{\hangindent=0.75em}p{2.75cm} X X X X X X} \toprule
     \multicolumn{2}{c}{Hyperparameters} & \multicolumn{2}{c}{Mortality} & \multicolumn{2}{c}{Primary CCS} & \multicolumn{2}{c}{All ICD-9} \\ \cmidrule{3-8}
     & & BOW & SHiP & BOW & SHiP & BOW & SHiP \\ \midrule
     Training & Learning rate & 0.00015 & 0.00011 & 0.00369 & 0.00067 & 0.00369 & 0.00048 \\
     & Batch size & 128 & 16 & 128 & 16 & 128 & 16 \\
     & Pretraining steps & -- & 30,000 & -- & 30,000 & -- & 40,000 \\
     & Gradient clip norm & 37.5 & 37.5 & 0.125 & 0.125 & 0.125 & 0.125 \\
     & Variational \mbox{vocabulary} dropout* & 0.001 & 0.229 & 0.273 & 0.396 & 0.273 & 0.273 \\ 
     & Bag length, hours & & & & & & \\ 
     & \quad \emph{Notes only} & 1 & 1 & 1 & 1 & 1 & 1 \\ 
     & \quad \emph{All features} & 1 & 1 & 8 & 8 & 8 & 8 \\ 
     & Maximum timesteps & & & & & & \\
     & \quad \emph{Notes only} & 1000 & 1000 & 1000 & 1000 & 1000 & 1000 \\ 
     & \quad \emph{All features} & 1000 & 1000 & 200 & 200 & 200 & 200 \\ \midrule 
     Record LSTM & Hidden units & \multicolumn{2}{c}{379} & \multicolumn{2}{c}{518} & \multicolumn{2}{c}{518} \\
     & Input dropout & \multicolumn{2}{c}{0.466} & \multicolumn{2}{c}{0.246} & \multicolumn{2}{c}{0.246} \\
     & Hidden dropout & \multicolumn{2}{c}{0.045} & \multicolumn{2}{c}{0.136} & \multicolumn{2}{c}{0.136} \\
     & Variational input dropout & \multicolumn{2}{c}{0.034} & \multicolumn{2}{c}{0.071} & \multicolumn{2}{c}{0.071} \\
     & Variational hidden dropout & \multicolumn{2}{c}{0.090} & \multicolumn{2}{c}{0.122} & \multicolumn{2}{c}{0.122} \\
     & Zoneout & \multicolumn{2}{c}{0.268} & \multicolumn{2}{c}{0.437} & \multicolumn{2}{c}{0.437} \\ \midrule
     Notes LSTM & Bidirectional? & -- & Yes & -- & Yes & -- & No \\
     & Hidden units & -- & 350 & -- & 325 & -- & 780 \\
     & Input dropout & -- & 0.052 & -- & 0.019 & -- & 0.340 \\
     & Hidden dropout & -- & 0.175 & -- & 0.391 & -- & 0.238 \\
     & Variational input dropout & -- & 0.176 & -- & 0.291 & -- & 0.156 \\
     & Variational hidden dropout & -- & 0.061 & -- & 0.085 & -- & 0.103 \\
     & Zoneout & -- & 0.312 & -- & 0.336 & -- & 0.387 \\ \bottomrule
  \end{tabularx}
  \caption*{* The variational vocabulary dropout rate is shared across all features. For baseline models with bigrams, we increased the dropout rate on the notes vocabulary only to 0.75.}
  \label{hparams} 
\end{table}

\newpage
\section{Additional experimental results}

\begin{table}[htbp]
  \caption{Comparison of different bagging lengths for all-feature hierarchical CCS and ICD9 models. Reporting mean (standard deviation) test set results over five runs from random initialization.} 
  \begin{tabularx}{\textwidth}{p{2.25cm} >{\hangindent=0.75em}p{3cm} X X X X} \toprule
  & & \multicolumn{2}{c}{Primary CCS} & \multicolumn{2}{c}{All ICD-9} \\ \cmidrule{3-6}
  & & Top-1 \mbox{Recall} & Top-5 \mbox{Recall} & AUPRC & AUROC, weighted \\ \midrule
  No pretraining & 1-hour bagging, \newline $t = 1000$ & 0.555 (0.020) & 0.812 (0.014)& 0.291 (0.010) & 0.869 (0.004) \\
  & 8-hour bagging, \newline $t = 200$ & 0.591 (0.008) & 0.833 (0.006) & 0.301 (0.004) & 0.868 (0.001) \\ \midrule
  SHiP & 1-hour bagging, \newline $t = 1000$ & 0.660 (0.004) & 0.887 (0.003) & 0.332 (0.016) & 0.889 (0.002) \\
  & 8-hour bagging, \newline $t = 200$ & 0.671 (0.004) & 0.890 (0.001) & 0.345 (0.005) & 0.889 (0.002) \\ \bottomrule
  \end{tabularx}
  \label{bagging} 
\end{table}

\begin{table}[htbp]
  \caption{Comparison of unidirectional vs. bidirectional notes LSTMs for SHiP models. Reporting validation set results for a single run.} 
  \begin{tabularx}{\textwidth}{p{2.5cm} X X X X X X} \toprule
  & \multicolumn{2}{c}{Mortality} & \multicolumn{2}{c}{Primary CCS} & \multicolumn{2}{c}{All ICD-9} \\ \cmidrule{2-7}
  & AUPRC & AUROC & Top-1 \mbox{Recall} & Top-5 \mbox{Recall} & AUPRC & AUROC, weighted \\ \midrule
  Unidirectional & 0.490 & 0.895 & 0.651 & 0.888 & 0.342 & 0.887 \\
  Bidirectional & 0.497 & 0.896 & 0.663 & 0.896 & 0.326 & 0.878 \\ \bottomrule
  \end{tabularx}
  \label{uni-vs-bi} 
\end{table}

\begin{table}[htbp]
  \caption{Comparison of pretraining time thresholds for all-feature SHiP mortality models. Reporting mean (standard deviation) test set results over five runs from random initialization.} 
  \begin{tabularx}{\textwidth}{p{6cm} X X} \toprule
  & AUPRC & AUROC \\ \midrule
  Pretrained to 24 hours & 0.478 (0.005) & 0.881 (0.001) \\
  Pretrained to discharge & 0.479 (0.007) & 0.882 (0.001) \\ \bottomrule
  \end{tabularx}
  \label{pretrain-time} 
\end{table}

\end{document}